\documentclass[10pt,a4paper]{article}
\usepackage{etoolbox}
\PassOptionsToPackage{table}{xcolor}
\usepackage{leapwing}
\usepackage{array,multirow,xspace}
\definecolor{cvprblue}{rgb}{0.21,0.49,0.74}
\definecolor{resultreference}{HTML}{EEF0F3}
\definecolor{resultours}{HTML}{E5F0FA}
\definecolor{methodblue}{HTML}{2166A5}
\newcommand{\nickname}{AoG\xspace}
\newcommand{\ourscell}[1]{\cellcolor{resultours}\textbf{#1}}

\setcitestyle{numbers,square,comma}

\renewcommand{\LWTitle}{Agent on Guard: Self-Improving Runtime Intervention for Robust Robotic Manipulation}
\renewcommand{\LWShortTitle}{AoG-RSI}
\renewcommand{\LWReportID}{Technical Report}
\renewcommand{\LWDate}{November 2026}
\renewcommand{\LWPartnerLogoHeight}{11mm}
\renewcommand{\LWPartnerLogos}{%
  \LWPartnerLogo[12mm]{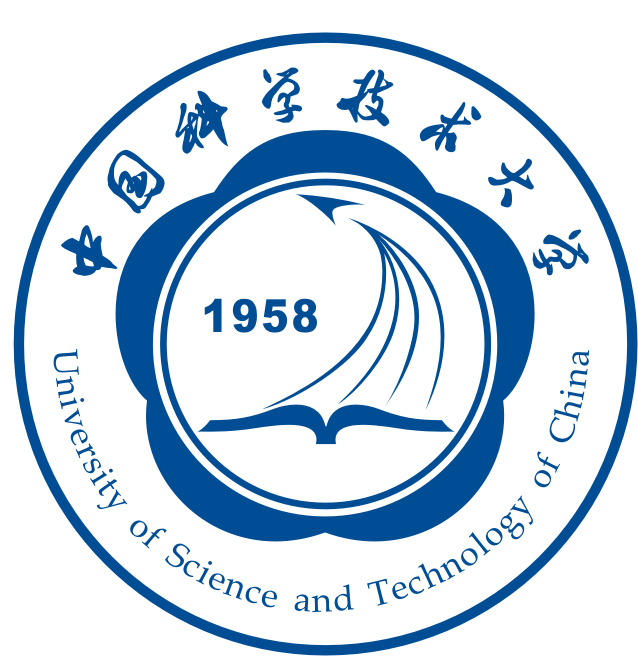}
}
\renewcommand{\LWAuthors}{%
  Gengming Cui\textsuperscript{1*}\quad
  Mingfei Han\textsuperscript{1,2,3*\textdagger}\quad
  Jiawei Zhang\textsuperscript{1}\quad
  Mozheng Liao\textsuperscript{1}\\[3pt]
  Yunkai Guo\textsuperscript{4}\quad
  Minghao Jin\textsuperscript{1}\quad
  Tingwen Du\textsuperscript{1}\quad
  Salman Khan\textsuperscript{3}\quad
  Xiaojun Chang\textsuperscript{1,2,3}\quad
  Zhihui Li\textsuperscript{1}}
\renewcommand{\LWAffiliations}{%
  \textsuperscript{1}Future Media Computing Lab, University of Science and Technology of China~
  \textsuperscript{2}LeapWing AI~
  \textsuperscript{3}MBZUAI~
  \textsuperscript{4}IA, CAS}
\renewcommand{\LWNotes}{%
  {\sffamily\color{LWMuted}
  \textsuperscript{*}Equal contribution\quad
  \textsuperscript{\textdagger}Project lead\quad
    }
}

\renewcommand{\LWAbstract}{
Robot foundation policies provide learned execution capabilities, while programming agents can reason about failures and construct recovery behaviors. Combining them effectively requires learning when intervention helps and when the policy can reliably resume. This challenge is inherently temporal: the physical world continues to evolve during reasoning, and a useful recovery may become infeasible before a decision takes effect. We introduce Agent on Guard (\nickname), a hierarchical framework that improves collaboration between a frozen base policy and a programming agent through physical experience. An evolving outer planner allocates control as the physical state evolves, aligning decisions with the physical clock with attention to observation freshness, remaining recovery windows, and the time required for reasoning, recovery, and subsequent execution. Coupled inner loops jointly refine intervention timing, recovery skills, and handoff conditions. Rather than judging recovery by local correction alone, \nickname uses downstream task progress, failure recurrence, final outcomes, and resource consumption to improve both recovery behaviors and scheduling. This turns temporary agent solutions into reusable collaboration capabilities that complement the frozen policy. On RoboDojo-Sim, \nickname achieves a success rate of 50.57\% and a task score of 57.26, exceeding the strongest evaluated baseline by 11.22 percentage points and 10.68 points, respectively. It also reduces end-to-end episode latency by 80.4\% and online token consumption by 85.1\% relative to the GPT-6 Astra agent.
}
\begin{document}
\makeLWTitle

\section{Introduction}
\label{sec:intro}

\begin{figure}[htbp]
    \centering
    \includegraphics[width=\textwidth]{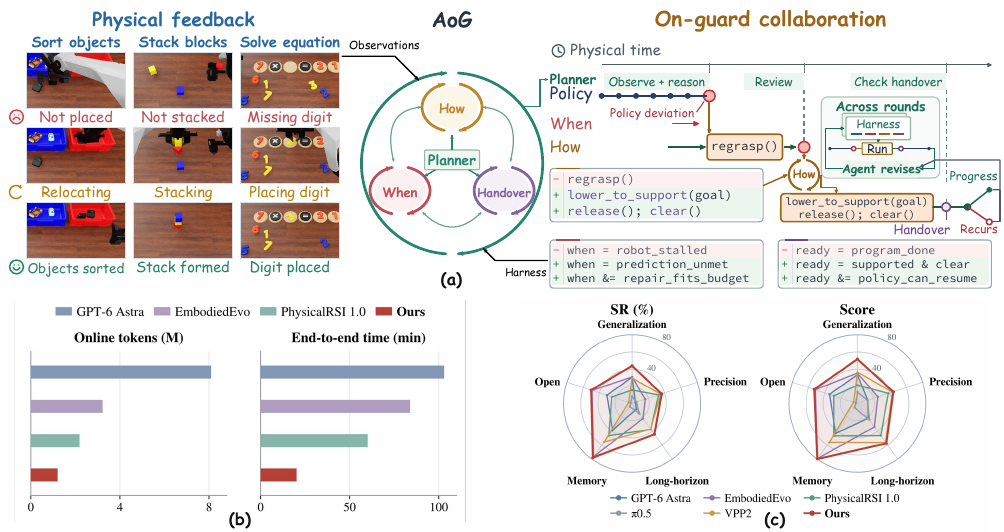}
    \caption{\textbf{Overview of \nickname.}
    (a) A planner coordinates agent reasoning with ongoing
    observation and frozen-policy execution under physical
    time constraints. Three coupled inner loops refine
    intervention timing, recovery programs, and policy resumption.
    Across RSI rounds, the agent revises planner guidance and
    inner-loop rules and programs from complete trajectories;
    subsequent executions use the updated harness.
    (b) Online execution efficiency on RoboDojo-Sim, comparing
    \nickname with agentic baselines in terms of token consumption
    and end-to-end latency per episode. Both metrics measure
    deployment costs, excluding RSI training costs.
    (c) Success rate and score across five task categories on
    RoboDojo-Sim, comparing \nickname with policy-only and
    agentic baselines.}
    \label{fig:streamingrsi_overview}

\end{figure}

Vision-language-action models (VLAs) and world action models (WAMs) learn reusable manipulation skills from large-scale data~\cite{kim2024openvla,intelligence2025pi05,ye2026dreamzero}. Adapting these skills to new task constraints, changing scenes, and execution errors remains challenging~\cite{chen2026robodojounifiedsimandrealbenchmark,shi2024yell}. Programming agents interpret goals, compose control programs, and revise them through execution feedback~\cite{liang2023codeaspolicies,huang2022innermonologue,fu2026capx}, while reliable grasping and fine contact control remain bottlenecks~\cite{li2026embodiedevoselfevolvingcodingagent,gao2026hexashell}. Physical recursive self-improvement (RSI) retains and revises programs, skills, and decision rules from physical experience~\cite{xiao2026enpire,song2026embodiedrsi}. These complementary strengths raise the question of how agent reasoning and program adaptation can extend the policy's capabilities while preserving its reliable execution.

Existing physical RSI systems remain constrained in when they can revise execution plans and control decisions as physical conditions change. Physical RSI 1.0~\cite{physicalrsi2026} selects a skill composition at the start of each episode and keeps it bound throughout execution. EmbodiedEvo~\cite{li2026embodiedevoselfevolvingcodingagent} and ASPIRE~\cite{lu2026aspire} organize observation, execution, and verification into discrete interaction rounds, with decision updates tied to round boundaries. A correction derived from an earlier observation may no longer suit the state in which it takes effect, while a locally successful repair may consume the time needed for subsequent execution. Control allocation must therefore evolve with recovery capabilities, accounting for when a recovery remains feasible and whether it leaves sufficient time for subsequent execution.

We introduce \textbf{\nickname}, a self-improving harness that enables a programming agent to monitor and assist a frozen robot policy throughout execution (Fig.~\ref{fig:streamingrsi_overview}a). An evolving planner allocates control according to the physical state and the time available for recovery and task completion. Coupled inner RSI loops refine when to intervene, how to recover, and when the policy can resume. The agent learns from what happens after recovery, using complete trajectories to refine both the planner and the inner loops. Recovery capabilities and control allocation thus evolve together: improved recovery changes the planner's choices, while the resulting execution provides new experience for refining both.
On RoboDojo-Sim~\cite{chen2026robodojounifiedsimandrealbenchmark}, \nickname achieves a success rate of 50.57\% and a task score of 57.26, outperforming the strongest evaluated baseline, WorldScape-R$^2$SI~\cite{manifold2026r2si}, by 11.22 percentage points and 10.68 points, respectively. Figure~\ref{fig:streamingrsi_overview}c presents the comparison
across five task categories. It also achieves the lowest end-to-end episode latency and online token consumption among the evaluated methods (Fig.~\ref{fig:streamingrsi_overview}b), reducing them by 80.4\% and 85.1\% relative to the GPT-6 Astra agent~\cite{zhang2026unexpectedrobotpolicyearly}.
Our contributions are threefold:
\begin{itemize}
    \item We introduce \textbf{\nickname}, a physical RSI framework that learns when to intervene, how to recover, and when to return control to a frozen policy under physical time constraints.
    
    \item We develop a mechanism that jointly evolves control allocation and recovery programs from complete trajectories, learning how the agent can complement a frozen policy and restore conditions for continued task execution.

    \item \nickname improves success rate and task score over the strongest evaluated baseline by 11.22 percentage points and 10.68 points, while reducing end-to-end episode latency by 80.4\% and online token consumption by 85.1\% relative to the GPT-6 Astra agent.
\end{itemize}

\section{Related Work}
\label{sec:related}

\subsection{Robot Foundation Policies and Programmatic Control}

Robot foundation policies learn reusable visuomotor behavior through large-scale pretraining~\cite{ghosh2024octo}, drawing on vision--language representations in VLA models~\cite{zitkovich2023rt2,kim2024openvla,intelligence2025pi06,intelligence2026pi07} and video prediction priors in world action models~\cite{guo2026vpp2,li2026lingbotva}.
Spatial action representations and language or multimodal conditioning support manipulation~\cite{zeng2021transporter,shridhar2022cliport,jiang2023vima}, while diffusion policies~\cite{chi2023diffusion} and action chunking~\cite{zhao2023act} model continuous action sequences for coordinated execution.
Programmatic approaches express behavior through executable logic: Code as Policies~\cite{liang2023codeaspolicies} composes perception and control APIs, while ProgPrompt~\cite{singh2022progprompt} grounds generated programs in available actions and objects.
VoxPoser~\cite{huang2023voxposer} and ReKep~\cite{huang2024rekep} further connect language reasoning to motion generation through 3D value maps and relational keypoint constraints, respectively.
Our focus is how programmatic recovery can complement a frozen policy and restore conditions for its continued execution.

\subsection{Runtime Orchestration of Robot Policies}

SayCan~\cite{ahn2022saycan} grounds skill selection in learned affordances, while Inner Monologue~\cite{huang2022innermonologue} incorporates execution feedback into replanning.
Related agent architectures interleave reasoning with action~\cite{yao2023react} or invoke deliberative planning alongside fast learned behavior~\cite{lin2023swiftsage}.
In robotics, uncertainty-aware help requests~\cite{ren2023knowno}, experience-based failure diagnosis~\cite{liu2023reflect}, and language corrections~\cite{sharma2022corrections} guide execution revision; corrective interactions can also improve subsequent decisions~\cite{shi2024yell,liang2024lmpc}.
Harness VLA~\cite{zhang2026harnessvla} combines frozen policies for contact-rich behavior with analytic primitives for non-contact motion.
HarnessWAM~\cite{gu2026harnesswam} couples frequent progress estimation with event-triggered task management, while DynaHarness~\cite{deng2026dynaharness} uses execution contracts to govern heterogeneous capabilities.
For execution timing, slow-fast visual--tactile policies~\cite{xue2025reactive} separate trajectory generation from rapid contact feedback, while real-time chunking (RTC)~\cite{black2025rtc} overlaps action generation with execution to accommodate inference latency.
Action-chunk continuity alone does not establish whether a recovery remains feasible when an agent decision takes effect. We study policy--agent control transfers under these physical time constraints.

\subsection{Physical Recursive Self-Improvement}

Agent self-improvement uses reflective memory~\cite{shinn2023reflexion} and iterative self-feedback~\cite{madaan2023selfrefine}, executable skill accumulation~\cite{wang2023voyager}, or reward-program optimization from training feedback~\cite{ma2023eureka}.
In physical environments, Physical RSI 1.0~\cite{physicalrsi2026}, ASPIRE~\cite{lu2026aspire}, and EmbodiedEvo~\cite{li2026embodiedevoselfevolvingcodingagent} accumulate skills and execution knowledge for subsequent tasks.
WorldScape-R$^2$SI~\cite{manifold2026r2si} improves skills and reactive-model weights through an outer recursion agent.
Beyond skill acquisition, Zetta~\cite{ding2026zetta} evolves runtime critics and recovery programs while retaining fixed high-level arbitration.
SERA~\cite{du2026sera} improves history-dependent planning by revising its task-context harness from rollouts, although planning updates remain tied to execution-round boundaries.
RoboDoctor~\cite{yu2026robodoctor} links action diagnosis and physical correction through reusable experience, while retaining the HYBRID/DIRECT mode selected after warmup.
We study the joint improvement of scheduling and recovery for continuous physical execution, using subsequent policy execution to assess their collaboration.

\section{Method}
\label{sec:method}

\begin{figure}[htbp]
    \centering
    \includegraphics[width=\textwidth]{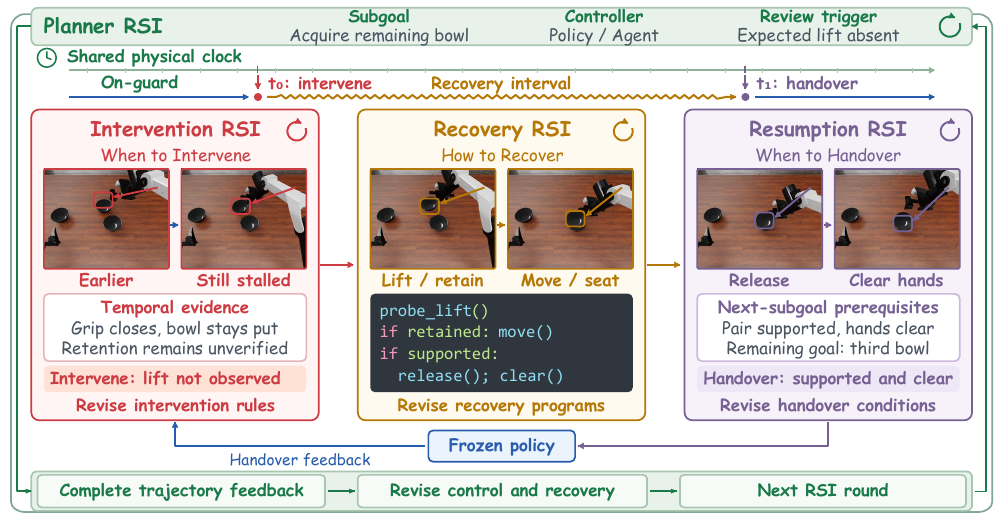}
    \caption{
    The planner coordinates agent observation and reasoning with
    ongoing frozen-policy execution on a shared physical clock.
    Temporal observations identify unmet object-level expectations
    and guide intervention at $t_0$.
    Recovery programs check grasp retention and object support,
    while handover at $t_1$ requires the physical prerequisites
    for the next policy subgoal.
    Subsequent policy execution reveals whether the recovered
    state supports continued task progress.
    Across RSI rounds, the agent uses this evidence within complete
    trajectories to jointly revise planner guidance, intervention
    rules, recovery programs, and handover conditions.
    }
    \label{fig:streamingrsi_method}

\end{figure}

\nickname{} organizes policy--agent collaboration into an outer planning loop and three inner loops for intervention, recovery, and resumption (Fig.~\ref{fig:streamingrsi_method}). The outer planner allocates control under physical time constraints; the inner loops refine when to intervene, how to recover, and when the policy can resume. Across rounds, the programming agent uses execution feedback to revise the guidance, decision rules, and recovery programs underlying both levels.

\subsection{Control Allocation under Runtime Constraints} 
\label{sec:physical_control}

Policy execution continues while the agent observes and reasons. Acting as the planner, the agent uses RGB observations, public robot state, action history, and previous handover feedback to choose continued policy execution, inspection, or recovery, without privileged simulator state. Its plan records subgoals, completion conditions, controller assignments, and estimated action requirements. The planner accounts for actions consumed during reasoning and the budget needed for recovery and subsequent policy execution. Loss of grasp or support can make a planned recovery infeasible even with budget remaining.

The host checks planner-specified monitors using public robot state and image-region or shape-matching signals after each confirmed policy action, allowing agent review before an action block finishes. At takeover, it clears queued policy actions; after handover, the policy generates fresh actions from updated observations.
For a recovery followed by handover in RSI round $k$, let $t_o$, $t_0$, and $t_1$ denote observation, takeover, and handover times. The physical state $s_t$ evolves as
\begin{equation}
\begin{aligned}
s_{t_0} &= \Phi_{\pi}\!\left(s_{t_o},\,t_0-t_o\right), \\
s_{t_1} &= \Phi_{r_k}\!\left(s_{t_0},\,t_1-t_0\right)
           \in \mathcal C_{\pi,k}.
\end{aligned}
\label{eq:physical_recovery}
\end{equation}
Here, $\Phi_c(s,\Delta t)$ denotes state evolution during execution by controller $c$ over elapsed time $\Delta t$, $\pi$ is the frozen policy, and $r_k$ is the recovery program. The set $\mathcal C_{\pi,k}$ contains states accepted by the current handover conditions for the policy's next subgoal.

\subsection{Recovery for Policy Resumption}
\label{sec:policy_resumption}

\noindent\textbf{Intervention.}
At each review, the agent compares expected object changes with current observations to choose continued execution, inspection, or repair. The decision accounts for evidence of progress, observable risk, and applicable failure experience. When the policy continues, the agent specifies the expected next effect, updates the monitoring conditions, and schedules the next review accordingly. Uncertain evidence can motivate further inspection, while previously learned failure conditions can support proactive takeover when matched by current observations.

\noindent\textbf{Recovery.}
Recovery addresses one physical bottleneck at a time, targeting the conditions required for the policy to continue its next subgoal. The agent works in a persistent Python session, selects applicable programs and experience, and adapts
the recovery code to the current scene. Before execution, the agent specifies the expected object changes. It executes short action sequences and checks their effects against these expectations, revising the program when execution does not produce the expected change. Recovery preserves completed subgoals and physical conditions needed for subsequent execution. The agent may complete the current subgoal or retain control through task completion when the policy cannot yet continue effectively, subject to the remaining budget.

\noindent\textbf{Resumption.}
The agent checks whether the observed object and robot states satisfy the prerequisites of the next policy subgoal and specifies the expected policy progress. After handover, the agent checks whether the repair remains effective, whether the policy advances the subgoal, and whether the original failure recurs. These observations determine whether policy execution continues or further inspection or recovery is needed.

\subsection{Evolving Policy--Agent Coordination}
\label{sec:coordination_evolution}

The programming agent jointly revises planner guidance, intervention rules and monitors, recovery programs, and handover conditions. It develops reusable procedures during physical execution, then consolidates them at episode termination using the complete trajectory and final outcome. This review connects each intervention to subsequent policy progress, failure recurrence, and action consumption, informing both program revisions and future control allocation.

The agent uses local recovery effects and subsequent policy execution to guide revisions. A program that fails to produce the expected object change requires changes to its procedure or applicability conditions. When the correction holds but the policy cannot advance, the agent instead reexamines the recovery endpoint, handover prerequisites, and whether to retain control longer. When intervention begins after necessary physical conditions have been lost, the agent revises monitoring conditions and review timing. Action consumption informs recovery limits and the budget reserved for policy continuation. The agent retains useful procedures with their applicable conditions and counterexamples from failed attempts.

Revised recovery programs $r_k$ change the time and physical conditions required to restore policy execution, while updated handover conditions $\mathcal C_{\pi,k}$ change which states are accepted for policy resumption. These changes guide the planner's controller assignments and intervention timing. The resulting control decisions expose recovery programs to different starting states, and their outcomes inform further revisions to both levels. Each round inherits the revised guidance and programs. Both models retain fixed weights throughout RSI. During evaluation, shared guidance and experience remain frozen; scene-specific code adaptation and feedback stay within the current episode.

\section{Experiments}
\label{sec:experiments}

\begin{table}[htbp]
\centering

\caption{\textbf{Main results on RoboDojo-Sim.}
We report Score and success rate (SR, \%) across five capability
dimensions and their macro-average (Overall).
Higher is better for both metrics.
Generalization combines the standard and randomized settings.}
\label{tab:robodojo_main}

\begingroup
\colorlet{methodblue}{black}

\setlength{\tabcolsep}{3.0pt}
\renewcommand{\arraystretch}{1.22}
\footnotesize

\resizebox{\textwidth}{!}{%
\begin{tabular}{
    ll
    >{\columncolor{resultreference}}c
    >{\columncolor{resultreference}}c
    *{10}{c}
}
\toprule
\multirow{2}{*}{\textbf{Method}}
& \multirow{2}{*}{\textbf{Type}}
& \multicolumn{2}{c}{
    \cellcolor{resultreference}\textbf{Overall}}
& \multicolumn{2}{c}{\textbf{Generalization}}
& \multicolumn{2}{c}{\textbf{Precision}}
& \multicolumn{2}{c}{\textbf{Long-Horizon}}
& \multicolumn{2}{c}{\textbf{Memory}}
& \multicolumn{2}{c}{\textbf{Open}} \\

\cmidrule(lr){3-4}
\cmidrule(lr){5-6}
\cmidrule(lr){7-8}
\cmidrule(lr){9-10}
\cmidrule(lr){11-12}
\cmidrule(lr){13-14}

&
& Score & SR (\%)
& Score & SR (\%)
& Score & SR (\%)
& Score & SR (\%)
& Score & SR (\%)
& Score & SR (\%) \\

\midrule

$\pi_{0.5}$~\cite{intelligence2025pi05}
& VLA
& 11.44 & 6.93
& 13.38 & 8.17
& 12.40 & 5.50
& 23.54 & 14.67
& 5.89 & 4.67
& 1.98 & 1.67 \\

Xiaomi-Robotics-1~\cite{team2026xiaomi}
& VLA
& 20.07 & 13.93
& 23.54 & 17.00
& 26.69 & 18.83
& 38.39 & 23.67
& 7.81 & 6.56
& 3.94 & 3.58 \\

GR00T-N1.7~\cite{nvidia2026groot17}
& VLA
& 2.85 & 1.31
& 2.16 & 1.22
& 2.54 & 0.67
& 8.30 & 3.58
& 1.06 & 0.89
& 0.18 & 0.17 \\

VPP2~\cite{guo2026vpp2}
& WAM
& 39.26 & 32.26
& 36.38 & 29.50
& 43.03 & 35.17
& 56.28 & 37.00
& 56.01 & 55.56
& 4.61 & 4.08 \\

GPT-6 Astra~\cite{zhang2026unexpectedrobotpolicyearly}
& Agent
& 28.97 & 22.48
& 33.36 & 30.50
& 12.65 & 4.00
& 21.45 & 8.25
& 43.04 & 38.67
& 34.36 & 31.00 \\

PhysicalRSI 1.0~\cite{physicalrsi2026}
& Agent + VLA
& 36.27 & 31.38
& 21.30 & 15.62
& 38.05 & 32.50
& 46.37 & 37.33
& 46.74 & 46.56
& 28.88 & 24.92 \\

EmbodiedEvo~\cite{li2026embodiedevoselfevolvingcodingagent}
& Agent
& 45.08 & 38.95
& 35.22 & 30.83
& 25.15 & 16.25
& 33.83 & 21.75
& 79.40 & 76.67
& 51.81 & 49.25 \\

WorldScape-R$^2$SI~\cite{manifold2026r2si}
& Agent + WAM
& 46.58 & 39.35
& 42.95 & 35.83
& 23.51 & 15.50
& 51.48 & 34.25
& 65.78 & 63.67
& 49.16 & 47.50 \\

\midrule

\rowcolor{resultours}
\textbf{\nickname (Ours)}
& \textbf{Agent + WAM}
& \ourscell{57.26} & \ourscell{50.57}
& \ourscell{51.60} & \ourscell{43.83}
& \ourscell{44.08} & \ourscell{36.25}
& \ourscell{57.65} & \ourscell{44.50}
& \ourscell{79.93} & \ourscell{78.00}
& \ourscell{53.04} & \ourscell{50.25} \\

\bottomrule
\end{tabular}%
}
\endgroup

\end{table}

\subsection{Experimental Setup}
\label{sec:experimental_setup}

\noindent\textbf{Benchmark and tasks.}
We evaluate \nickname on RoboDojo-Sim~\cite{chen2026robodojounifiedsimandrealbenchmark},
a simulation benchmark for bimanual manipulation with the ARX X5
robot. The benchmark comprises 42 tasks across five capability
dimensions: Generalization (12 tasks), Precision (8 tasks),
Long-Horizon (8 tasks), Memory (6 tasks), and Open (8 tasks).
These dimensions assess robustness to scene variations, precise
physical interaction, multi-step execution, history-dependent
decision-making, and open-vocabulary instruction following,
respectively. 
We additionally evaluate \nickname on the 50-task
multi-task benchmark of RoboCasa365~\cite{robocasa365},
comprising 18 Atomic-Seen, 16 Composite-Seen, and
16 Composite-Unseen tasks.

\noindent\textbf{Baselines and configurations.}
We compare \nickname with representative VLA, WAM, and agent-based
methods. VLA baselines include
$\pi_{0.5}$~\cite{intelligence2025pi05},
Xiaomi-Robotics-1~\cite{team2026xiaomi}, and
GR00T-N1.7~\cite{nvidia2026groot17}.
We include VPP2~\cite{guo2026vpp2} as a WAM baseline.
Agent-based baselines comprise
GPT-6 Astra~\cite{zhang2026unexpectedrobotpolicyearly} and
EmbodiedEvo~\cite{li2026embodiedevoselfevolvingcodingagent}.
We additionally compare with
PhysicalRSI 1.0~\cite{physicalrsi2026} and WorldScape-R$^2$SI~\cite{manifold2026r2si},
which combine an agent with a VLA and a WAM policy, respectively.
On RoboDojo-Sim, \nickname pairs VPP2 with
GPT-6 Astra~\cite{openai2026gpt6astra} (xhigh).
On RoboCasa365, it pairs Xiaomi-Robotics-1 with either
GPT-6 Astra (xhigh) or GPT-6.1 Sol~\cite{openai2026gpt61sol} (xhigh).
We evolve the harness separately for each benchmark and policy--agent configuration over five cumulative RSI rounds.

\noindent\textbf{Evaluation protocol and metrics.}
We evaluate \nickname on all 42 RoboDojo-Sim tasks
with 50 episodes per task, including 25 standard and
25 randomized episodes for each Generalization task.
We retain the benchmark's action budgets and success
criteria.
To limit evaluation cost, we use a development evaluation set for ablations and analyses, covering all 42 tasks with five scenes per task (210 episodes per configuration). Each Generalization task includes three standard and two randomized scenes.
The evolved harness and accumulated experience remain
frozen during evaluation, with no experience updates
across test episodes or environment resets within an episode.
We report Score (0--100), which measures task completion
including partial progress, and success rate (SR, \%),
the percentage of episodes achieving full task success.
Overall is the unweighted mean of the five
capability-level scores and success rates.
For RoboCasa365, we evaluate 50 episodes per task and
report success rates for each split and the overall
mean across all 50 tasks.

\subsection{Main Results}
\label{sec:main_results}

\noindent\textbf{Overall performance.}
\nickname achieves 50.57\% SR and 57.26 Score on RoboDojo-Sim, exceeding the strongest evaluated baseline by 11.22 percentage points and 10.68 points, respectively (Table~\ref{tab:robodojo_main}). VPP2 performs better than EmbodiedEvo on Precision and Long-Horizon, while EmbodiedEvo is stronger on Open and Memory. \nickname outperforms both across these categories, suggesting that the policy's standalone performance understates the reach of its learned manipulation skills. Agent-guided planning and recovery help make these skills effective in situations the policy struggles to handle independently.
On RoboCasa365, \nickname improves Xiaomi-Robotics-1 across all three splits with both GPT-6 Astra and GPT-6.1 Sol (Table~\ref{tab:robocasa365}). Both configurations yield larger gains on composite tasks, particularly Composite-Unseen. These tasks require individual manipulation skills to remain effective across successive stages, where an earlier execution error can prevent later actions from succeeding. The concentration of gains on these tasks suggests that agent-guided planning and recovery help the policy carry its existing skills through longer and unfamiliar task compositions.

\begin{table}[htbp]
    \centering
    \caption{
    Success rates (\%) on RoboCasa365 with Xiaomi-Robotics-1
    as the frozen base policy and two programming agents.
    Overall is the mean success rate across all 50 tasks.
    }
    \label{tab:robocasa365}
    \footnotesize
    \setlength{\tabcolsep}{3pt}
    \renewcommand{\arraystretch}{1.12}
    \begin{tabular}{
        >{\raggedright\arraybackslash}p{\dimexpr0.36\linewidth-2\tabcolsep\relax}
        *{4}{>{\centering\arraybackslash}p{\dimexpr0.16\linewidth-2\tabcolsep\relax}}
    }
        \toprule
        Method
        & \shortstack{Atomic\\Seen}
        & \scalebox{0.9}{\shortstack{Composite\\Seen}}
        & \scalebox{0.9}{\shortstack{Composite\\Unseen}}
        & Overall \\
        \midrule
        Xiaomi-Robotics-1~\cite{team2026xiaomi}
            & 80.2 & 57.1 & 32.1 & 57.4 \\
        \midrule
        \rowcolor{resultours}
        Ours (GPT-6 Astra)
            & \textbf{88.1} & \textbf{68.8}
            & \textbf{49.6} & \textbf{69.6} \\
        \rowcolor{resultours}
        Ours (GPT-6.1 Sol)
            & 84.0 & 63.9 & 40.8 & 63.7 \\
        \bottomrule
    \end{tabular}

\end{table}

\begin{figure}[htbp]
    \centering
    \includegraphics[width=\textwidth]{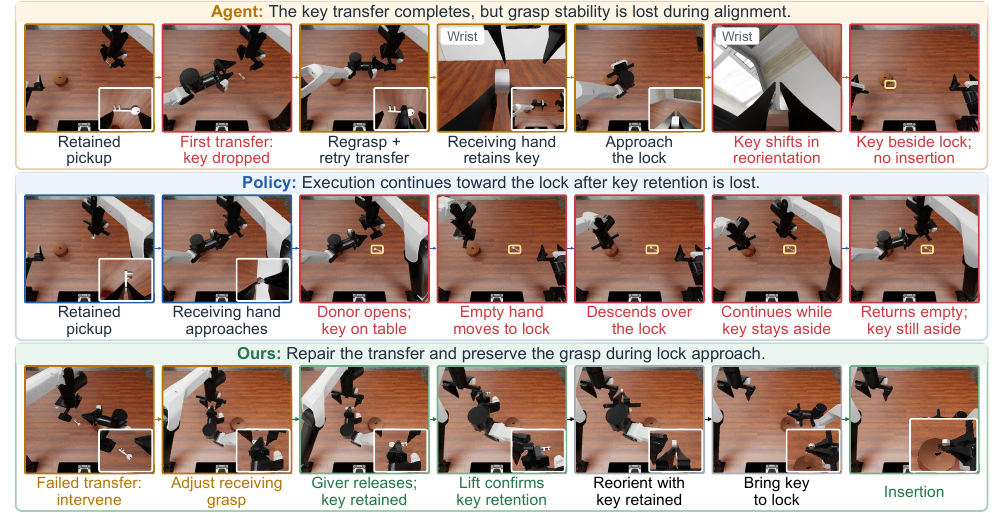}
    \caption{
    \textbf{Qualitative comparison on bimanual key insertion.}
    Sequences progress from left to right.
    Top: the standalone agent retries a failed transfer and
    brings the key toward the lock, but loses stable grasp
    during reorientation.
    Middle: the standalone policy drops the key during transfer
    yet continues toward the lock with an empty gripper.
    Bottom: \nickname corrects the receiving grasp after a
    failed transfer, retains the key through reorientation
    and approach, and completes insertion.
    Insets and wrist views show key retention and contact details.
    }
    \label{fig:case_insert_key}

\end{figure}

\begin{figure}[htbp]
    \centering
    \includegraphics[width=.5\linewidth]
        {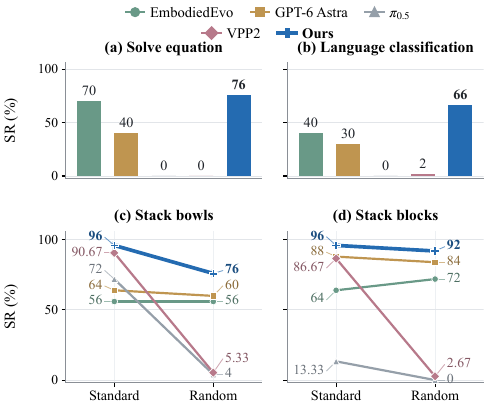}
    \caption{Success rates (SR, \%) on RoboDojo-Sim
    Open (a,b) and Generalization (c,d) tasks.
    (a) Solve Equation requires selecting and placing
    the correct missing number or operator.
    (b) Language Classification requires sorting unseen
    objects into baskets according to language-specified
    category assignments.
    (c) Stack Bowls requires stacking three bowls upright.
    (d) Stack Blocks requires stably stacking three
    differently textured blocks.
    Lines in (c,d) connect results under standard and
    randomized settings.}
    \label{fig:robodojo_generalization}

\end{figure}

\begin{figure}[htbp]
    \centering
    \includegraphics[width=.64\linewidth]
        {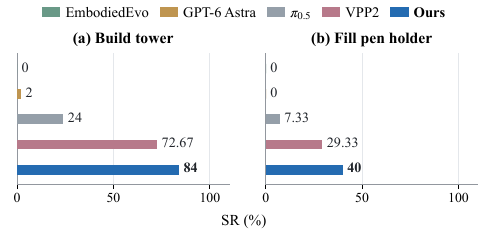}
    \caption{Success rates (SR, \%) on RoboDojo-Sim tasks
    requiring fine control.
    (a) Build Tower requires precisely positioning wooden
    blocks and boards to keep successive layers centered
    and stable.
    (b) Fill Pen Holder requires one arm to hold the holder
    upright while the other sequentially inserts four pens.}
    \label{fig:robodojo_execution}
\end{figure}

\noindent\textbf{Complementary strengths of policies and agents.}
Agents outperform standalone policies on the evaluated Open tasks, which require interpreting new goals and composing actions under semantic and language constraints (Fig.~\ref{fig:robodojo_generalization} a, b). On Generalization tasks, the standard-to-randomized comparisons reveal a separate limitation: policy competence on familiar scenes does not consistently transfer to changed visual and spatial conditions, even when the task goal remains unchanged (Fig.~\ref{fig:robodojo_generalization} c, d). Conversely, policies outperform agents on the selected tasks requiring precise contact, sustained object support, and coordinated execution (Fig.~\ref{fig:robodojo_execution}). Across these settings, \nickname improves on the stronger standalone controller, combining gains on Open tasks and randomized scenes with improvements in fine-control tasks.
Figure~\ref{fig:case_insert_key} reveals two distinct requirements for collaboration. The policy continues after losing a prerequisite for insertion, whereas the agent restores key possession without securing a grasp that survives reorientation. Intervention must detect when task prerequisites are lost, and recovery must restore them in a form that supports subsequent motion. \nickname retains the key through reorientation and insertion, illustrating how downstream execution tests whether the transfer was effectively repaired.

\noindent\textbf{Execution efficiency.}
\nickname achieves the highest SR and Score with the lowest online token consumption and end-to-end latency on matched scenes (Table~\ref{tab:execution_efficiency}). Efficient recovery also requires recognizing when further agent reasoning is unnecessary: once the policy can resume, continued agent control adds computation to steps the policy can already perform. Learned handover conditions limit this overhead, while reusable programs reduce repeated recovery development. Overlapping reasoning with policy execution further reduces waiting time.

\begin{table}[htbp]
    \centering
    \caption{
    Execution efficiency on the RoboDojo-Sim development evaluation set.
    Tokens and end-to-end latency are averaged per episode,
    excluding RSI training costs.
    }
    \label{tab:execution_efficiency}
    \footnotesize
    \setlength{\tabcolsep}{3pt}
    \renewcommand{\arraystretch}{1.08}
    \begin{tabular}{
        p{\dimexpr.32\columnwidth-2\tabcolsep\relax}
        >{\centering\arraybackslash}p{\dimexpr.14\columnwidth-2\tabcolsep\relax}
        >{\centering\arraybackslash}p{\dimexpr.14\columnwidth-2\tabcolsep\relax}
        >{\centering\arraybackslash}p{\dimexpr.20\columnwidth-2\tabcolsep\relax}
        >{\centering\arraybackslash}p{\dimexpr.20\columnwidth-2\tabcolsep\relax}
    }
        \toprule
        \textbf{Method} & \textbf{SR (\%)}
               & \textbf{Score}
               & \textbf{Tokens (M)}
               & \textbf{Time (min)} \\

        \midrule
        GPT-6 Astra~\cite{zhang2026unexpectedrobotpolicyearly}     & 22.50 & 29.03 & 8.096 & 103.0 \\
        EmbodiedEvo~\cite{li2026embodiedevoselfevolvingcodingagent}     & 37.83 & 44.52 & 3.224 & 83.9  \\
        PhysicalRSI 1.0~\cite{physicalrsi2026} & 31.50 & 36.37 & 2.188 & 60.1  \\
        \midrule
        \rowcolor{resultours}
        \textbf{\nickname} & \ourscell{54.17}
                        & \ourscell{58.92}
                        & \ourscell{1.204}
                        & \ourscell{20.2} \\
        \bottomrule
    \end{tabular}
\end{table}

\subsection{Ablation Study}
\label{sec:ablation}

\begin{figure}[htbp]
    \centering
    \includegraphics[width=.6\linewidth]
        {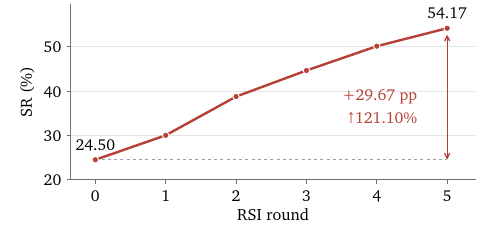}
    \caption{
Success rate across cumulative RSI rounds on the RoboDojo-Sim development evaluation set, with the base policy frozen.
Round 0 denotes the initial harness.
}
    \label{fig:rsi_evolution}
\end{figure}

\noindent\textbf{Evolution across RSI rounds.}
\nickname improves SR from 24.50\% to 54.17\% over five RSI rounds (Fig.~\ref{fig:rsi_evolution}). Execution feedback aligns recovery behaviors with the conditions needed for subsequent policy progress, while planner updates adapt control allocation to these evolving capabilities. The initial harness underperforms the standalone policy because of unnecessary interventions, ineffective recovery actions, and poorly timed handovers. These results show that effective collaboration requires learning both how to assist the policy when it struggles and how to preserve useful execution that intervention can disrupt.

\begin{figure}[htbp]
    \centering
    \includegraphics[width=.64\linewidth]
        {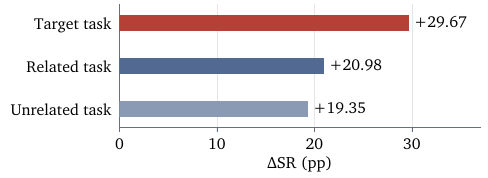}
    \caption{
Cross-task transfer of evolved harnesses.
The same target tasks are evaluated using frozen harnesses
from the target task, a related task, or an unrelated task.
$\Delta$SR measures improvement over the initial harness
in percentage points.
}
    \label{fig:rsi_transfer}
\end{figure}

\noindent\textbf{Evolved harnesses transfer across tasks.}
Harnesses from other tasks improve performance on the same target evaluations (Fig.~\ref{fig:rsi_transfer}).
Related-task and unrelated-task harnesses increase SR over the initial harness by 20.98 and 19.35 percentage
points, respectively. The substantial gains from unrelated tasks suggest that RSI captures reusable collaboration knowledge beyond task-specific solutions, while the modest advantage of related tasks indicates some benefit from matching execution requirements. Target-task harnesses achieve the largest gain,
highlighting the additional value of adapting intervention, recovery, and handover to the task's specific constraints.
Appendix~C describes how related and unrelated source tasks are selected and lists the source--target task pairs.

\begin{table}[htbp]
    \centering
    \caption{
Ablations of RSI scope and execution feedback on the RoboDojo-Sim development evaluation set.
RSI scope determines which components can improve;
all others retain their initial implementations.
Local feedback excludes post-handover outcomes from updates.
Full \nickname jointly evolves the planner and three
inner loops using complete trajectories.
}
    \label{tab:ablations}
    \footnotesize
    \setlength{\tabcolsep}{3pt}
    \renewcommand{\arraystretch}{1.08}
    \begin{tabular}{
        p{\dimexpr.20\columnwidth-2\tabcolsep\relax}
        p{\dimexpr.45\columnwidth-2\tabcolsep\relax}
        >{\centering\arraybackslash}p{\dimexpr.175\columnwidth-2\tabcolsep\relax}
        >{\centering\arraybackslash}p{\dimexpr.175\columnwidth-2\tabcolsep\relax}
    }
        \toprule
        \textbf{Ablation} & \textbf{Configuration}
                 & \textbf{SR (\%)} & \textbf{Score}  \\
        \midrule
        \multirow{5}{*}{RSI scope}
        & No RSI                  & 24.50 & 29.77 \\
        & Recovery only           & 31.83 & 36.39 \\
        & Inner loops             & 46.83 & 52.53 \\
        & \quad w/o intervention  & 38.17 & 45.52 \\
        & \quad w/o resumption    & 36.00 & 42.38 \\
        \midrule
        Feedback
        & Local only              & 49.00 & 54.23 \\
        \midrule
        \rowcolor{resultours}
        \multicolumn{2}{>{\columncolor{resultours}}l}{\textbf{Full \nickname}}
        & \ourscell{54.17} & \ourscell{58.92} \\
        \bottomrule
    \end{tabular}
\end{table}

\noindent\textbf{Evolving recovery and control allocation.}
Recovery programs and control allocation benefit from joint evolution (Table~\ref{tab:ablations}). Updating intervention and resumption conditions substantially improves on recovery-only evolution, while freezing either update reduces the performance of inner-loop RSI. The larger drop from freezing resumption indicates that recovery improvements are harder to exploit when handover conditions remain fixed. The additional gain from planner evolution supports revising task-level control assignments as the agent's recovery capabilities and the policy's opportunities to resume change.

\noindent\textbf{Learning from post-handover execution.}
Feedback after handover is necessary to distinguish local repair from useful recovery. Full-trajectory feedback improves both SR and Score over local feedback alone (Table~\ref{tab:ablations}). Local correction establishes whether the agent achieved its immediate objective, while subsequent policy execution reveals whether that correction enables further task progress. Learning from this distinction helps revise recovery targets and handover conditions that appear adequate locally but fail during continuation.

\subsection{Real-World Experiments}
\label{sec:real_world}

\begin{figure}[htbp]
    \centering
    \includegraphics[width=\textwidth]{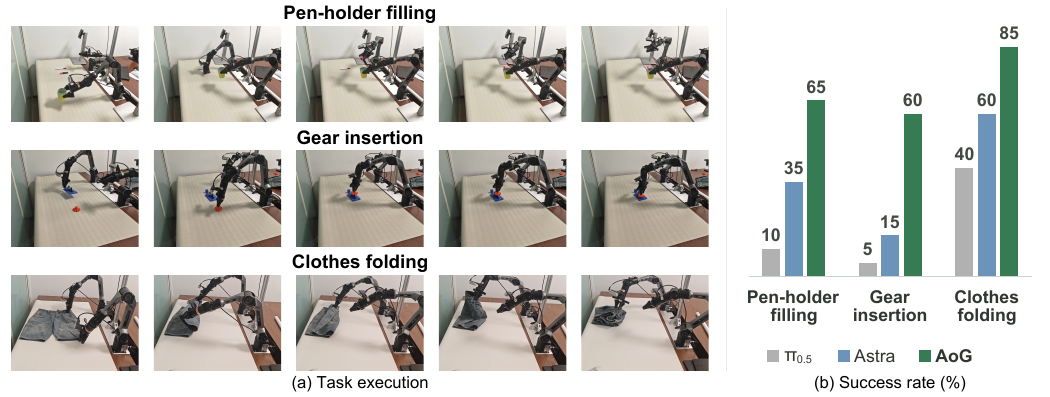}
    \caption{Real-world evaluation. (a) Selected execution stages for pen-holder filling, gear insertion, and clothes folding. (b) Task success rates over 20 trials per method, comparing \nickname{} with agent-only GPT-6 Astra and fine-tuned $\pi_{0.5}$.}
    \label{fig:realworld_overview}
\end{figure}

\noindent\textbf{Real-world evaluation.}
We use a dual-arm AgileX PiPER with three RGB-D cameras, one global and two wrist-mounted. For each task, we fine-tune $\pi_{0.5}$ on 20 real-robot demonstrations and freeze its weights. After four RSI rounds, we freeze the evolved harness and accumulated collaboration experience for testing. We compare \nickname{}, agent-only GPT-6 Astra, and fine-tuned $\pi_{0.5}$ on 20 independent test trials per task and method (Fig.~\ref{fig:realworld_overview}). All methods use the same observation interface, robot controller, and action budgets, and are evaluated under matched initial scene conditions. Success requires full task completion without human intervention.

\noindent\textbf{Pen-holder filling.}
One arm holds the pen holder upright while the other grasps, aligns, and inserts the pens. With the policy weights unchanged, \nickname{} increases success from 10\% to 65\%. Insertion stalled despite continued arm motion, prompting the agent to correct the pen's position and orientation. The supporting grasp remained intact throughout the correction, keeping the holder stable. After handover, the policy resumed insertion from the corrected alignment without disrupting the support provided by the other arm.

\noindent\textbf{Gear insertion.}
\nickname{} reaches 60\% success in gear assembly. The right arm centers the base while the left inserts the loose gear onto its post to mesh with the mounted gear. Repeated positioning attempts failed to seat the gear: centering it over the post was insufficient without the orientation needed to engage the neighboring gear. The agent corrected the alignment while the other arm maintained the base position. After handover, the policy seated the gear and completed engagement without displacing the base.

\noindent\textbf{Clothes folding.}
\nickname{} achieves 85\% success on clothes folding, where both standalone controllers already complete many trials. The robot folds the cloth through successive bimanual grasps, reorientations, and placements, each of which changes the grasp points available for the next step. When the cloth configuration prevented the next planned grasp, the agent restored access to the grasp point while preserving completed folds. The policy then continued the remaining sequence. Unlike rigid-object alignment, this correction depended on the folds already formed and had to preserve the cloth arrangement needed for subsequent steps.

\section{Conclusion}
\label{sec:conclusion}
We presented \nickname, a hierarchical RSI framework that improves collaboration between a frozen robot policy and a programming agent. Its evolving planner allocates control as physical execution continues, accounting for reasoning delays and finite recovery windows. Coupled inner loops refine intervention timing, recovery programs, and handover conditions through feedback from subsequent execution. On RoboDojo-Sim, \nickname improves task performance while reducing online token consumption and end-to-end latency. The framework learns how to recover together with when to intervene and return control, aligning collaboration with the physical clock.

\section*{LLM Usage Statement}
OpenAI ChatGPT was used solely for language editing of the manuscript. Large language models were also used as programming agents in the experiments, as described in the methodology and experimental setup. The authors take full responsibility for the final content and all AI-assisted materials.

\clearpage
\begingroup
\small
\bibliographystyle{ieeenat_fullname}
\bibliography{reference}

\begin{thebibliography}{54}
\providecommand{\natexlab}[1]{#1}
\providecommand{\url}[1]{\texttt{#1}}
\expandafter\ifx\csname urlstyle\endcsname\relax
  \providecommand{\doi}[1]{doi: #1}\else
  \providecommand{\doi}{doi: \begingroup \urlstyle{rm}\Url}\fi

\bibitem[Ahn et~al.(2022)Ahn, Brohan, Brown, Chebotar, Cortes, David, Finn, Fu, Gopalakrishnan, Hausman, Herzog, Ho, Hsu, Ibarz, Ichter, Irpan, Jang, Jauregui~Ruano, Jeffrey, Jesmonth, Joshi, Julian, Kalashnikov, Kuang, Lee, Levine, Lu, Luu, Parada, Pastor, Quiambao, Rao, Rettinghouse, Reyes, Sermanet, Sievers, Tan, Toshev, Vanhoucke, Xia, Xiao, Xu, Xu, Yan, and Zeng]{ahn2022saycan}
Michael Ahn, Anthony Brohan, Noah Brown, Yevgen Chebotar, Omar Cortes, Byron David, Chelsea Finn, Chuyuan Fu, Keerthana Gopalakrishnan, Karol Hausman, Alex Herzog, Daniel Ho, Jasmine Hsu, Julian Ibarz, Brian Ichter, Alex Irpan, Eric Jang, Rosario Jauregui~Ruano, Kyle Jeffrey, Sally Jesmonth, Nikhil~J Joshi, Ryan Julian, Dmitry Kalashnikov, Yuheng Kuang, Kuang-Huei Lee, Sergey Levine, Yao Lu, Linda Luu, Carolina Parada, Peter Pastor, Jornell Quiambao, Kanishka Rao, Jarek Rettinghouse, Diego Reyes, Pierre Sermanet, Nicolas Sievers, Clayton Tan, Alexander Toshev, Vincent Vanhoucke, Fei Xia, Ted Xiao, Peng Xu, Sichun Xu, Mengyuan Yan, and Andy Zeng.
\newblock Do as i can, not as i say: Grounding language in robotic affordances.
\newblock \emph{arXiv preprint arXiv:2204.01691}, 2022.

\bibitem[Black et~al.(2025)Black, Galliker, and Levine]{black2025rtc}
Kevin Black, Manuel~Y. Galliker, and Sergey Levine.
\newblock Real-time execution of action chunking flow policies, 2025.

\bibitem[Chen et~al.(2026)Chen, Chen, Li, Tang, Su, Lu, Wan, Chen, Liu, Yan, Su, Dou, Wang, Zhang, Liu, Qin, Liang, Wu, Lin, Lin, Wang, He, Wu, Wu, Zhou, Lei, Yu, Ji, Jin, Lin, Li, Xiong, Xu, Li, Chai, Xie, Wang, Mu, Dong, Matusik, Ding, Ding, Luo, and Tomizuka]{chen2026robodojounifiedsimandrealbenchmark}
Tianxing Chen, Yue Chen, Zixuan Li, Junyuan Tang, Kailun Su, Haoran Lu, Weijie Wan, Baijun Chen, Songling Liu, Haowen Yan, Honghao Su, Zhiyang Dou, Kaixuan Wang, Dandan Zhang, Yunze Liu, Yan Qin, Qiwei Liang, Qiwei Wu, Zijian Lin, Wenwei Lin, Yuran Wang, Minghua He, Tianshu Wu, Ruihai Wu, Jingquan Zhou, Kai-Chong Lei, Haibao Yu, Yuanfeng Ji, Weiyang Jin, Guanyu Lin, Xiaofan Li, Qi Xiong, Renjing Xu, Zhongyu Li, Wenhao Chai, Enze Xie, Ziwei Wang, Yao Mu, Hao Dong, Wojciech Matusik, Mingyu Ding, Wenbo Ding, Ping Luo, and Masayoshi Tomizuka.
\newblock {RoboDojo}: A unified sim-and-real benchmark for comprehensive evaluation of generalist robot manipulation policies, 2026.

\bibitem[Chi et~al.(2023)Chi, Feng, Du, Xu, Cousineau, Burchfiel, and Song]{chi2023diffusion}
Cheng Chi, Siyuan Feng, Yilun Du, Zhenjia Xu, Eric Cousineau, Benjamin Burchfiel, and Shuran Song.
\newblock {Diffusion Policy: Visuomotor Policy Learning via Action Diffusion}.
\newblock In \emph{Proceedings of Robotics: Science and Systems}, 2023.

\bibitem[Deng et~al.(2026)Deng, Liu, Zhang, Yan, Cao, and Wang]{deng2026dynaharness}
Haoyuan Deng, Jiebin Liu, Tengxiao Zhang, Langning Yan, Hongye Cao, and Ziwei Wang.
\newblock {DynaHarness}: A dynamic physical harness for self-evolving robot agents.
\newblock \emph{arXiv preprint arXiv:2609.40306}, 2026.

\bibitem[Ding et~al.(2026)Ding, Mi, Huang, Wang, Zhang, Hao, Chen, Li, Zheng, Guo, Wang, Li, Wu, Liu, and Cao]{ding2026zetta}
Xin Ding, Liang Mi, Mingzhe Huang, Zixuan Wang, Chao Zhang, Zixu Hao, Fu Chen, Xiangyu Li, Yikai Zheng, Yaoyu Guo, Weijun Wang, Kun Li, Hao Wu, Yunxin Liu, and Ting Cao.
\newblock {Zetta} $\zeta$: An efficient closed-loop embodied harness for self-evolving physical intelligence.
\newblock \emph{arXiv preprint arXiv:2608.16590}, 2026.

\bibitem[Du et~al.(2026)Du, Guo, Zhang, Wang, Dong, Gao, Han, Lu, Yu, Shen, Chang, and Li]{du2026sera}
Tingwen Du, Haoyu Guo, Hanxue Zhang, Xuanxu Wang, Junting Dong, Chao Gao, Mingfei Han, Tao Lu, Mulin Yu, Chunhua Shen, Xiaojun Chang, and Zhihui Li.
\newblock From {VLM} planners to robotic agents: Self-evolving harnesses for history-dependent manipulation, 2026.

\bibitem[Fu et~al.(2026)Fu, Yu, El-Refai, Kou, Xue, Huang, Xiao, Wang, Niu, Li, Shi, Wu, Sastry, Zhu, Goldberg, and Fan]{fu2026capx}
Letian Fu, Justin Yu, Karim El-Refai, Ethan Kou, Haoru Xue, Huang Huang, Wenli Xiao, Guanzhi Wang, Dantong Niu, Fei-Fei Li, Guanya Shi, Jiajun Wu, Shankar Sastry, Yuke Zhu, Ken Goldberg, and Linxi~"Jim" Fan.
\newblock {CaP-X}: A framework for benchmarking and improving coding agents for robot manipulation, 2026.

\bibitem[Gao et~al.(2026)Gao, Zeng, Zhou, Ge, Shangguan, Wang, Wan, Zheng, Zhu, Li, Wu, Wang, He, Jia, Wu, and He]{gao2026hexashell}
Hongcheng Gao, Jianshu Zeng, Alex Zhou, Shijia Ge, Xuan Shangguan, Yazhe Wang, Yexing Wan, Zelin Zheng, Jay Zhu, Xiping Li, Di Wu, Yongli Wang, Lingyu He, Zhiqi Jia, Sihang Wu, and Xiao He.
\newblock {HexaShell}: A universal harness for robot coding agents in simulation and the real world.
\newblock \emph{arXiv preprint arXiv:2610.23479}, 2026.

\bibitem[Ghosh et~al.(2024)Ghosh, Walke, Pertsch, Black, Mees, Dasari, Hejna, Kreiman, Xu, Luo, Tan, Chen, Vuong, Xiao, Sanketi, Sadigh, Finn, and Levine]{ghosh2024octo}
Dibya Ghosh, Homer~Rich Walke, Karl Pertsch, Kevin Black, Oier Mees, Sudeep Dasari, Joey Hejna, Tobias Kreiman, Charles Xu, Jianlan Luo, You~Liang Tan, Lawrence~Yunliang Chen, Quan Vuong, Ted Xiao, Pannag~R. Sanketi, Dorsa Sadigh, Chelsea Finn, and Sergey Levine.
\newblock {Octo: An Open-Source Generalist Robot Policy}.
\newblock In \emph{Proceedings of Robotics: Science and Systems}, 2024.

\bibitem[Gu et~al.(2026)Gu, Zhu, Lin, Zhu, Chen, Wang, Yuan, Yang, Zhao, Li, Su, and Wang]{gu2026harnesswam}
Zhaopeng Gu, Bingke Zhu, Tianxi Lin, Guibo Zhu, Yingying Chen, Kai Wang, Tingyu Yuan, Fan Yang, Chaoyang Zhao, Zhaowen Li, Peng Su, and Jinqiao Wang.
\newblock {HarnessWAM}: Bridging prediction and deliberation in world action models.
\newblock \emph{arXiv preprint arXiv:2608.09516}, 2026.

\bibitem[Guo et~al.(2026)Guo, Yan, Zhong, Zhang, Yang, Lu, Chen, Wang, Deng, Yang, Yuan, Liu, Ban, Zhu, Liu, Feng, Li, and Chen]{guo2026vpp2}
Yanjiang Guo, Haodong Yan, Zhide Zhong, Zhongru Zhang, Qingyuan Yang, Qingzhou Lu, Xiaoyu Chen, Yen-Jen Wang, Shuying Deng, Chenghan Yang, Puzhen Yuan, Chenxin Liu, Tun Ban, Xiang Zhu, Yichen Liu, Kun Feng, Haoang Li, and Jianyu Chen.
\newblock Video prediction policy 2: Predict better, act better.
\newblock \emph{arXiv preprint arXiv:2610.10270}, 2026.

\bibitem[Huang et~al.(2022)Huang, Xia, Xiao, Chan, Liang, Florence, Zeng, Tompson, Mordatch, Chebotar, Sermanet, Brown, Jackson, Luu, Levine, Hausman, and Ichter]{huang2022innermonologue}
Wenlong Huang, Fei Xia, Ted Xiao, Harris Chan, Jacky Liang, Pete Florence, Andy Zeng, Jonathan Tompson, Igor Mordatch, Yevgen Chebotar, Pierre Sermanet, Noah Brown, Tomas Jackson, Linda Luu, Sergey Levine, Karol Hausman, and Brian Ichter.
\newblock Inner monologue: Embodied reasoning through planning with language models.
\newblock \emph{arXiv preprint arXiv:2207.05608}, 2022.

\bibitem[Huang et~al.(2023)Huang, Wang, Zhang, Li, Wu, and Fei-Fei]{huang2023voxposer}
Wenlong Huang, Chen Wang, Ruohan Zhang, Yunzhu Li, Jiajun Wu, and Li Fei-Fei.
\newblock {VoxPoser}: Composable {3D} value maps for robotic manipulation with language models.
\newblock \emph{arXiv preprint arXiv:2307.05973}, 2023.

\bibitem[Huang et~al.(2024)Huang, Wang, Li, Zhang, and Fei-Fei]{huang2024rekep}
Wenlong Huang, Chen Wang, Yunzhu Li, Ruohan Zhang, and Li Fei-Fei.
\newblock {ReKep}: Spatio-temporal reasoning of relational keypoint constraints for robotic manipulation.
\newblock \emph{arXiv preprint arXiv:2409.01652}, 2024.

\bibitem[Jiang et~al.(2023)Jiang, Gupta, Zhang, Wang, Dou, Chen, Fei-Fei, Anandkumar, Zhu, and Fan]{jiang2023vima}
Yunfan Jiang, Agrim Gupta, Zichen Zhang, Guanzhi Wang, Yongqiang Dou, Yanjun Chen, Li Fei-Fei, Anima Anandkumar, Yuke Zhu, and Linxi Fan.
\newblock {VIMA}: Robot manipulation with multimodal prompts.
\newblock In \emph{Proceedings of the 40th International Conference on Machine Learning}, pages 14975--15022. PMLR, 2023.

\bibitem[Kim et~al.(2024)Kim, Pertsch, Karamcheti, Xiao, Balakrishna, Nair, Rafailov, Foster, Lam, Sanketi, Vuong, Kollar, Burchfiel, Tedrake, Sadigh, Levine, Liang, and Finn]{kim2024openvla}
Moo~Jin Kim, Karl Pertsch, Siddharth Karamcheti, Ted Xiao, Ashwin Balakrishna, Suraj Nair, Rafael Rafailov, Ethan Foster, Grace Lam, Pannag Sanketi, Quan Vuong, Thomas Kollar, Benjamin Burchfiel, Russ Tedrake, Dorsa Sadigh, Sergey Levine, Percy Liang, and Chelsea Finn.
\newblock {OpenVLA}: An open-source vision-language-action model.
\newblock \emph{arXiv preprint arXiv:2406.09246}, 2024.

\bibitem[Li et~al.(2026{\natexlab{a}})Li, Zhang, Luo, Yang, Wang, Han, Yu, Gao, Xue, Zhu, Shen, and Xu]{li2026lingbotva}
Lin Li, Qihang Zhang, Yiming Luo, Shuai Yang, Ruilin Wang, Fei Han, Mingrui Yu, Zelin Gao, Nan Xue, Xing Zhu, Yujun Shen, and Yinghao Xu.
\newblock Causal world modeling for robot control, 2026{\natexlab{a}}.

\bibitem[Li et~al.(2026{\natexlab{b}})Li, Chi, Zheng, Liu, Wang, Gao, Jiang, Hong, Xiao, Yang, and Na]{li2026embodiedevoselfevolvingcodingagent}
Wenyi Li, Yizhe Chi, Youjie Zheng, Chunyu Liu, Xiaoqiu Wang, Mingju Gao, Zheng Jiang, Deyao Hong, Calvin Xiao, Kaisen Yang, and Qinhuai Na.
\newblock {EmbodiedEvo}: Self-evolving coding agent for physical intelligence, 2026{\natexlab{b}}.

\bibitem[Liang et~al.(2023)Liang, Huang, Xia, Xu, Hausman, Ichter, Florence, and Zeng]{liang2023codeaspolicies}
Jacky Liang, Wenlong Huang, Fei Xia, Peng Xu, Karol Hausman, Brian Ichter, Pete Florence, and Andy Zeng.
\newblock Code as policies: Language model programs for embodied control, 2023.

\bibitem[Liang et~al.(2024)Liang, Xia, Yu, Zeng, Arenas, Attarian, Bauza, Bennice, Bewley, Dostmohamed, Fu, Gileadi, Giustina, Gopalakrishnan, Hasenclever, Humplik, Hsu, Joshi, Jyenis, Kew, Kirmani, Lee, Lee, Michaely, Moore, Oslund, Rao, Ren, Tabanpour, Vuong, Wahid, Xiao, Xu, Zhuang, Xu, Frey, Caluwaerts, Zhang, Ichter, Tompson, Takayama, Vanhoucke, Shafran, Mataric, Sadigh, Heess, Rao, Stewart, Tan, and Parada]{liang2024lmpc}
Jacky Liang, Fei Xia, Wenhao Yu, Andy Zeng, Montserrat~Gonzalez Arenas, Maria Attarian, Maria Bauza, Matthew Bennice, Alex Bewley, Adil Dostmohamed, Chuyuan~Kelly Fu, Nimrod Gileadi, Marissa Giustina, Keerthana Gopalakrishnan, Leonard Hasenclever, Jan Humplik, Jasmine Hsu, Nikhil Joshi, Ben Jyenis, Chase Kew, Sean Kirmani, Tsang-Wei~Edward Lee, Kuang-Huei Lee, Assaf~Hurwitz Michaely, Joss Moore, Ken Oslund, Dushyant Rao, Allen Ren, Baruch Tabanpour, Quan Vuong, Ayzaan Wahid, Ted Xiao, Ying Xu, Vincent Zhuang, Peng Xu, Erik Frey, Ken Caluwaerts, Tingnan Zhang, Brian Ichter, Jonathan Tompson, Leila Takayama, Vincent Vanhoucke, Izhak Shafran, Maja Mataric, Dorsa Sadigh, Nicolas Heess, Kanishka Rao, Nik Stewart, Jie Tan, and Carolina Parada.
\newblock {Learning to Learn Faster from Human Feedback with Language Model Predictive Control}.
\newblock In \emph{Proceedings of Robotics: Science and Systems}, 2024.

\bibitem[Lin et~al.(2023)Lin, Fu, Yang, Brahman, Huang, Bhagavatula, Ammanabrolu, Choi, and Ren]{lin2023swiftsage}
Bill~Yuchen Lin, Yicheng Fu, Karina Yang, Faeze Brahman, Shiyu Huang, Chandra Bhagavatula, Prithviraj Ammanabrolu, Yejin Choi, and Xiang Ren.
\newblock {SwiftSage: A Generative Agent with Fast and Slow Thinking for Complex Interactive Tasks}.
\newblock In \emph{Advances in Neural Information Processing Systems}, 2023.

\bibitem[Liu et~al.(2023)Liu, Bahety, and Song]{liu2023reflect}
Zeyi Liu, Arpit Bahety, and Shuran Song.
\newblock Reflect: Summarizing robot experiences for failure explanation and correction.
\newblock In \emph{Proceedings of The 7th Conference on Robot Learning}, pages 3468--3484. PMLR, 2023.

\bibitem[Llontop and Vadrevu(2026)]{nvidia2026groot17}
Edith Llontop and Kalyan Vadrevu.
\newblock {NVIDIA Isaac GR00T N1.7}: Open reasoning {VLA} model for humanoid robots.
\newblock NVIDIA, Hugging Face blog, 2026.

\bibitem[Lu et~al.(2026)Lu, Wu, Kou, Lin, Fu, Xiao, Mandlekar, Xu, Shi, Goldberg, Chen, Chowdhury, Kautz, Zhu, Fan, and Wang]{lu2026aspire}
Runyu Lu, Yubo Wu, Ethan Kou, Haotian Lin, Letian Fu, Wenli Xiao, Ajay Mandlekar, Yinzhen Xu, Guanya Shi, Ken Goldberg, Ang Chen, Mosharaf Chowdhury, Jan Kautz, Yuke Zhu, Linxi~"Jim" Fan, and Guanzhi Wang.
\newblock {ASPIRE}: Agentic {/Skills} discovery for robotics.
\newblock \emph{arXiv preprint arXiv:2607.00272}, 2026.

\bibitem[Ma et~al.(2024)Ma, Liang, Wang, Huang, Bastani, Jayaraman, Zhu, Fan, and Anandkumar]{ma2023eureka}
Yecheng~Jason Ma, William Liang, Guanzhi Wang, De-An Huang, Osbert Bastani, Dinesh Jayaraman, Yuke Zhu, Linxi Fan, and Anima Anandkumar.
\newblock {Eureka}: Human-level reward design via coding large language models.
\newblock In \emph{International Conference on Learning Representations}, 2024.

\bibitem[Madaan et~al.(2023)Madaan, Tandon, Gupta, Hallinan, Gao, Wiegreffe, Alon, Dziri, Prabhumoye, Yang, Gupta, Majumder, Hermann, Welleck, Yazdanbakhsh, and Clark]{madaan2023selfrefine}
Aman Madaan, Niket Tandon, Prakhar Gupta, Skyler Hallinan, Luyu Gao, Sarah Wiegreffe, Uri Alon, Nouha Dziri, Shrimai Prabhumoye, Yiming Yang, Shashank Gupta, Bodhisattwa~Prasad Majumder, Katherine Hermann, Sean Welleck, Amir Yazdanbakhsh, and Peter Clark.
\newblock {Self-Refine: Iterative Refinement with Self-Feedback}.
\newblock In \emph{Advances in Neural Information Processing Systems}, 2023.

\bibitem[{Manifold AI Team}(2026)]{manifold2026r2si}
{Manifold AI Team}.
\newblock {R$^2$SI}: Recursive robot self-improvement via a bi-level agentic system with agent--model co-evolution.
\newblock Technical report, Manifold AI, 2026.

\bibitem[{MMLab \& PhysicalRSI contributors}(2026)]{physicalrsi2026}
{MMLab \& PhysicalRSI contributors}.
\newblock {Physical RSI 1.0}: Recursive self-harness for scaling embodied skills.
\newblock Project website and open-source software, 2026.

\bibitem[Nasiriany et~al.(2026)Nasiriany, Nasiriany, Maddukuri, and Zhu]{robocasa365}
Soroush Nasiriany, Sepehr Nasiriany, Abhiram Maddukuri, and Yuke Zhu.
\newblock {RoboCasa365}: A large-scale simulation framework for training and benchmarking generalist robots.
\newblock In \emph{International Conference on Learning Representations (ICLR)}, 2026.

\bibitem[{OpenAI}(2026{\natexlab{a}})]{openai2026gpt61sol}
{OpenAI}.
\newblock Introducing {GPT-6.1 Sol}, 2026{\natexlab{a}}.

\bibitem[{OpenAI}(2026{\natexlab{b}})]{openai2026gpt6astra}
{OpenAI}.
\newblock {GPT-6 Astra}: A new generation of intelligence, 2026{\natexlab{b}}.

\bibitem[{Physical Intelligence} et~al.(2025{\natexlab{a}}){Physical Intelligence}, Amin, Aniceto, Balakrishna, Black, Conley, Connors, Darpinian, Dhabalia, DiCarlo, Driess, Equi, Esmail, Fang, Finn, Glossop, Godden, Goryachev, Groom, Hancock, Hausman, Hussein, Ichter, Jakubczak, Jen, Jones, Katz, Ke, Kuchi, Lamb, LeBlanc, Levine, Li-Bell, Lu, Mano, Mothukuri, Nair, Pertsch, Ren, Sharma, Shi, Smith, Springenberg, Stachowicz, Stoeckle, Swerdlow, Tanner, Torne, Vuong, Walling, Wang, Williams, Yoo, Yu, Zhilinsky, and Zhou]{intelligence2025pi06}
{Physical Intelligence}, Ali Amin, Raichelle Aniceto, Ashwin Balakrishna, Kevin Black, Ken Conley, Grace Connors, James Darpinian, Karan Dhabalia, Jared DiCarlo, Danny Driess, Michael Equi, Adnan Esmail, Yunhao Fang, Chelsea Finn, Catherine Glossop, Thomas Godden, Ivan Goryachev, Lachy Groom, Hunter Hancock, Karol Hausman, Gashon Hussein, Brian Ichter, Szymon Jakubczak, Rowan Jen, Tim Jones, Ben Katz, Liyiming Ke, Chandra Kuchi, Marinda Lamb, Devin LeBlanc, Sergey Levine, Adrian Li-Bell, Yao Lu, Vishnu Mano, Mohith Mothukuri, Suraj Nair, Karl Pertsch, Allen~Z. Ren, Charvi Sharma, Lucy~Xiaoyang Shi, Laura Smith, Jost~Tobias Springenberg, Kyle Stachowicz, Will Stoeckle, Alex Swerdlow, James Tanner, Marcel Torne, Quan Vuong, Anna Walling, Haohuan Wang, Blake Williams, Sukwon Yoo, Lili Yu, Ury Zhilinsky, and Zhiyuan Zhou.
\newblock {$\pi^{*}_{0.6}$}: A {VLA} that learns from experience, 2025{\natexlab{a}}.

\bibitem[{Physical Intelligence} et~al.(2025{\natexlab{b}}){Physical Intelligence}, Black, Brown, Darpinian, Dhabalia, Driess, Esmail, Equi, Finn, Fusai, Galliker, Ghosh, Groom, Hausman, Ichter, Jakubczak, Jones, Ke, LeBlanc, Levine, Li-Bell, Mothukuri, Nair, Pertsch, Ren, Shi, Smith, Springenberg, Stachowicz, Tanner, Vuong, Walke, Walling, Wang, Yu, and Zhilinsky]{intelligence2025pi05}
{Physical Intelligence}, Kevin Black, Noah Brown, James Darpinian, Karan Dhabalia, Danny Driess, Adnan Esmail, Michael Equi, Chelsea Finn, Niccolo Fusai, Manuel~Y. Galliker, Dibya Ghosh, Lachy Groom, Karol Hausman, Brian Ichter, Szymon Jakubczak, Tim Jones, Liyiming Ke, Devin LeBlanc, Sergey Levine, Adrian Li-Bell, Mohith Mothukuri, Suraj Nair, Karl Pertsch, Allen~Z. Ren, Lucy~Xiaoyang Shi, Laura Smith, Jost~Tobias Springenberg, Kyle Stachowicz, James Tanner, Quan Vuong, Homer Walke, Anna Walling, Haohuan Wang, Lili Yu, and Ury Zhilinsky.
\newblock {$\pi_{0.5}$}: A vision-language-action model with open-world generalization, 2025{\natexlab{b}}.

\bibitem[{Physical Intelligence} et~al.(2026){Physical Intelligence}, Ai, Amin, Aniceto, Balakrishna, Balke, Black, Bokinsky, Cao, Charbonnier, Choudhary, Collins, Conley, Connors, Darpinian, Dhabalia, Dhaka, DiCarlo, Driess, Equi, Esmail, Fang, Finn, Glossop, Godden, Goryachev, Groom, Habeeb, Hancock, Hausman, Hussein, Hwang, Ichter, Jacobsen, Jakubczak, Jen, Jones, Kammerer, Katz, Ke, Khadikov, Kuchi, Lamb, LeBlanc, LeCount, Levine, Li, Li-Bell, Lialin, Liang, Lim, Lu, Luo, Mano, Marwaha, Mongush, Murphy, Nair, Patterson, Pertsch, Ren, Schelske, Sharma, Shi, Shi, Smith, Springenberg, Stachowicz, Stoeckle, Tang, Tanner, Tekeste, Torne, Vedder, Vuong, Walling, Wang, Wang, Wang, Whalen, Whitmore, Williams, Xu, Yoo, Yu, Zhang, Zhang, and Zhilinsky]{intelligence2026pi07}
{Physical Intelligence}, Bo Ai, Ali Amin, Raichelle Aniceto, Ashwin Balakrishna, Greg Balke, Kevin Black, George Bokinsky, Shihao Cao, Thomas Charbonnier, Vedant Choudhary, Foster Collins, Ken Conley, Grace Connors, James Darpinian, Karan Dhabalia, Maitrayee Dhaka, Jared DiCarlo, Danny Driess, Michael Equi, Adnan Esmail, Yunhao Fang, Chelsea Finn, Catherine Glossop, Thomas Godden, Ivan Goryachev, Lachlan Groom, Haroun Habeeb, Hunter Hancock, Karol Hausman, Gashon Hussein, Victor Hwang, Brian Ichter, Connor Jacobsen, Szymon Jakubczak, Rowan Jen, Tim Jones, Gregg Kammerer, Ben Katz, Liyiming Ke, Mairbek Khadikov, Chandra Kuchi, Marinda Lamb, Devin LeBlanc, Brendon LeCount, Sergey Levine, Xinyu Li, Adrian Li-Bell, Vladislav Lialin, Zhonglin Liang, Wallace Lim, Yao Lu, Enyu Luo, Vishnu Mano, Nandan Marwaha, Aikys Mongush, Liam Murphy, Suraj Nair, Tyler Patterson, Karl Pertsch, Allen~Z. Ren, Gavin Schelske, Charvi Sharma, Baifeng Shi, Lucy~Xiaoyang Shi, Laura Smith, Jost~Tobias Springenberg, Kyle Stachowicz, Will
  Stoeckle, Jiaming Tang, Jimmy Tanner, Shalom Tekeste, Marcel Torne, Kyle Vedder, Quan Vuong, Anna Walling, Haohuan Wang, Jason Wang, XuDong Wang, Chris Whalen, Samuel Whitmore, Blake Williams, Charles Xu, Sukwon Yoo, Lili Yu, Wuming Zhang, Zhuoyang Zhang, and Ury Zhilinsky.
\newblock {$\pi_{0.7}$}: A steerable generalist robotic foundation model with emergent capabilities, 2026.

\bibitem[Ren et~al.(2023)Ren, Dixit, Bodrova, Singh, Tu, Brown, Xu, Takayama, Xia, Varley, Xu, Sadigh, Zeng, and Majumdar]{ren2023knowno}
Allen~Z. Ren, Anushri Dixit, Alexandra Bodrova, Sumeet Singh, Stephen Tu, Noah Brown, Peng Xu, Leila Takayama, Fei Xia, Jake Varley, Zhenjia Xu, Dorsa Sadigh, Andy Zeng, and Anirudha Majumdar.
\newblock Robots that ask for help: Uncertainty alignment for large language model planners.
\newblock In \emph{Proceedings of The 7th Conference on Robot Learning}, pages 661--682. PMLR, 2023.

\bibitem[Sharma et~al.(2022)Sharma, Sundaralingam, Blukis, Paxton, Hermans, Torralba, Andreas, and Fox]{sharma2022corrections}
Pratyusha Sharma, Balakumar Sundaralingam, Valts Blukis, Chris Paxton, Tucker Hermans, Antonio Torralba, Jacob Andreas, and Dieter Fox.
\newblock {Correcting Robot Plans with Natural Language Feedback}.
\newblock In \emph{Proceedings of Robotics: Science and Systems}, 2022.

\bibitem[Shi et~al.(2024)Shi, Hu, Zhao, Sharma, Pertsch, Luo, Levine, and Finn]{shi2024yell}
Lucy~Xiaoyang Shi, Zheyuan Hu, Tony~Z. Zhao, Archit Sharma, Karl Pertsch, Jianlan Luo, Sergey Levine, and Chelsea Finn.
\newblock Yell at your robot: Improving on-the-fly from language corrections.
\newblock In \emph{Proceedings of Robotics: Science and Systems}, Delft, Netherlands, 2024.

\bibitem[Shinn et~al.(2023)Shinn, Cassano, Gopinath, Narasimhan, and Yao]{shinn2023reflexion}
Noah Shinn, Federico Cassano, Ashwin Gopinath, Karthik Narasimhan, and Shunyu Yao.
\newblock {Reflexion}: Language agents with verbal reinforcement learning.
\newblock In \emph{Advances in Neural Information Processing Systems}, 2023.

\bibitem[Shridhar et~al.(2022)Shridhar, Manuelli, and Fox]{shridhar2022cliport}
Mohit Shridhar, Lucas Manuelli, and Dieter Fox.
\newblock Cliport: What and where pathways for robotic manipulation.
\newblock In \emph{Proceedings of the 5th Conference on Robot Learning}, pages 894--906. PMLR, 2022.

\bibitem[Singh et~al.(2022)Singh, Blukis, Mousavian, Goyal, Xu, Tremblay, Fox, Thomason, and Garg]{singh2022progprompt}
Ishika Singh, Valts Blukis, Arsalan Mousavian, Ankit Goyal, Danfei Xu, Jonathan Tremblay, Dieter Fox, Jesse Thomason, and Animesh Garg.
\newblock {ProgPrompt}: Generating situated robot task plans using large language models.
\newblock \emph{arXiv preprint arXiv:2209.11302}, 2022.

\bibitem[Song et~al.(2026)Song, Liang, Fu, Wang, Yang, and Liu]{song2026embodiedrsi}
Python Song, Zhixuan Liang, Kelsey Fu, Mengdi Wang, Junfeng Yang, and Shilong Liu.
\newblock {EmbodiedRSI}: Active continual robot learning through hypothesis-guided co-evolution.
\newblock \emph{arXiv preprint arXiv:2610.10498}, 2026.

\bibitem[Wang et~al.(2023)Wang, Xie, Jiang, Mandlekar, Xiao, Zhu, Fan, and Anandkumar]{wang2023voyager}
Guanzhi Wang, Yuqi Xie, Yunfan Jiang, Ajay Mandlekar, Chaowei Xiao, Yuke Zhu, Linxi Fan, and Anima Anandkumar.
\newblock {Voyager}: An open-ended embodied agent with large language models.
\newblock \emph{arXiv preprint arXiv:2305.16291}, 2023.

\bibitem[Xiao et~al.(2026)Xiao, Xie, Zhang, Lin, Fu, Xue, Lu, Yang, Dai, Wang, Wu, Wang, Sastry, Goldberg, Fan, Zhu, and Shi]{xiao2026enpire}
Wenli Xiao, Jia Xie, Tonghe Zhang, Haotian Lin, Letian~"Max" Fu, Haoru Xue, Jalen Lu, Yi Yang, Cunxi Dai, Zi Wang, Jimmy Wu, Guanzhi Wang, S.~Shankar Sastry, Ken Goldberg, Linxi~"Jim" Fan, Yuke Zhu, and Guanya Shi.
\newblock {ENPIRE}: Agentic robot policy self-improvement in the real world, 2026.

\bibitem[{Xiaomi Robotics Team} et~al.(2026){Xiaomi Robotics Team}, Guo, Jin, Li, Li, Li, Liu, Peng, Qin, Su, Sun, Sun, Suo, Wang, Wang, Wu, Xia, Zhang, Zhao, Chen, Chen, He, Li, Li, Li, Qu, Song, Xiang, Xie, Xu, Ye, Ye, Zhao, and Zhou]{team2026xiaomi}
{Xiaomi Robotics Team}, Jun Guo, Piaopiao Jin, Jason Li, Peiyan Li, Yingyan Li, Futeng Liu, Wanli Peng, Optimus Qin, Yifei Su, Nan Sun, Qiao Sun, Runze Suo, Heyun Wang, Yunhong Wang, Rujie Wu, Caoyu Xia, Lina Zhang, Jack Zhao, Guoliang Chen, Wenlong Chen, Xinze He, Bin Li, Qing Li, Zhuorong Li, Heng Qu, Wenxuan Song, Diyun Xiang, Yifan Xie, Peiran Xu, Hangjun Ye, Wen Ye, Han Zhao, and Quanyun Zhou.
\newblock {Xiaomi-Robotics-1}: Scaling vision-language-action models with over {100K} hours of real-world trajectories.
\newblock \emph{arXiv preprint arXiv:2607.15330}, 2026.

\bibitem[Xue et~al.(2025)Xue, Ren, Chen, Zhang, Yuan, Gu, Xu, and Lu]{xue2025reactive}
Han Xue, Jieji Ren, Wendi Chen, Gu Zhang, Fang Yuan, Guoying Gu, Huazhe Xu, and Cewu Lu.
\newblock {Reactive Diffusion Policy: Slow-Fast Visual-Tactile Policy Learning for Contact-Rich Manipulation}.
\newblock In \emph{Proceedings of Robotics: Science and Systems}, 2025.

\bibitem[Yao et~al.(2023)Yao, Zhao, Yu, Du, Shafran, Narasimhan, and Cao]{yao2023react}
Shunyu Yao, Jeffrey Zhao, Dian Yu, Nan Du, Izhak Shafran, Karthik Narasimhan, and Yuan Cao.
\newblock {ReAct: Synergizing Reasoning and Acting in Language Models}.
\newblock In \emph{International Conference on Learning Representations}, 2023.

\bibitem[Ye et~al.(2026)Ye, Ge, Zheng, Gao, Yu, Kurian, Indupuru, Tan, Zhu, Xiang, Malik, Lee, Liang, Ranawaka, Gu, Xu, Wang, Hu, Narayan, Bjorck, Wang, Kim, Niu, Zheng, Xie, Wu, Wang, Julian, Xu, Du, Chebotar, Reed, Kautz, Zhu, Fan, and Jang]{ye2026dreamzero}
Seonghyeon Ye, Yunhao Ge, Kaiyuan Zheng, Shenyuan Gao, Sihyun Yu, George Kurian, Suneel Indupuru, You~Liang Tan, Chuning Zhu, Jiannan Xiang, Ayaan Malik, Kyungmin Lee, William Liang, Nadun Ranawaka, Jiasheng Gu, Yinzhen Xu, Guanzhi Wang, Fengyuan Hu, Avnish Narayan, Johan Bjorck, Jing Wang, Gwanghyun Kim, Dantong Niu, Ruijie Zheng, Yuqi Xie, Jimmy Wu, Qi Wang, Ryan Julian, Danfei Xu, Yilun Du, Yevgen Chebotar, Scott Reed, Jan Kautz, Yuke Zhu, Linxi~"Jim" Fan, and Joel Jang.
\newblock World action models are zero-shot policies, 2026.

\bibitem[Yu et~al.(2026)Yu, Li, Li, Yan, Peng, Zhong, Xin, Yao, Tian, and Li]{yu2026robodoctor}
Heyang Yu, Haomin Li, Sihang Li, Jixin Yan, Shiyun Peng, Wanqi Zhong, Tianyu Xin, Yuan Yao, Xiaoyu Tian, and Yiming Li.
\newblock {RoboDoctor}: Training-free recursive self-improvement for robot policies, 2026.

\bibitem[Zeng et~al.(2021)Zeng, Florence, Tompson, Welker, Chien, Attarian, Armstrong, Krasin, Duong, Sindhwani, and Lee]{zeng2021transporter}
Andy Zeng, Pete Florence, Jonathan Tompson, Stefan Welker, Jonathan Chien, Maria Attarian, Travis Armstrong, Ivan Krasin, Dan Duong, Vikas Sindhwani, and Johnny Lee.
\newblock Transporter networks: Rearranging the visual world for robotic manipulation.
\newblock In \emph{Proceedings of the 2020 Conference on Robot Learning}, pages 726--747. PMLR, 2021.

\bibitem[Zhang et~al.(2026{\natexlab{a}})Zhang, Wang, Ouyang, Huang, Li, Su, Jin, Chai, Liang, Dou, Chen, and Chen]{zhang2026unexpectedrobotpolicyearly}
Wenbo Zhang, Kaixuan Wang, Yutao Ouyang, Xiaoyu Huang, Liyang Li, Kailun Su, Weiyang Jin, Wenhao Chai, Haotian Liang, Zhiyang Dou, Yue Chen, and Tianxing Chen.
\newblock An unexpected robot policy: Early evaluations of {GPT-6 Astra} on {RoboDojo} and beyond, 2026{\natexlab{a}}.

\bibitem[Zhang et~al.(2026{\natexlab{b}})Zhang, Zhang, Gao, Li, Liu, Nie, Zhu, Qiu, Yan, Liu, Tang, Rao, Fang, Wei, Wang, Ding, and Yu]{zhang2026harnessvla}
Yixian Zhang, Huanming Zhang, Feng Gao, Xiao Li, Zhihao Liu, Yi Nie, Chunyang Zhu, Jiaxing Qiu, Yuchen Yan, Jiyuan Liu, Wenhao Tang, Jiaji Rao, Zhengru Fang, Changxu Wei, Yu Wang, Wenbo Ding, and Chao Yu.
\newblock Harness {VLA}: Steering frozen {VLA}s into reliable manipulation primitives via memory-guided agents.
\newblock \emph{arXiv preprint arXiv:2607.08448}, 2026{\natexlab{b}}.

\bibitem[Zhao et~al.(2023)Zhao, Kumar, Levine, and Finn]{zhao2023act}
Tony~Z. Zhao, Vikash Kumar, Sergey Levine, and Chelsea Finn.
\newblock {Learning Fine-Grained Bimanual Manipulation with Low-Cost Hardware}.
\newblock In \emph{Proceedings of Robotics: Science and Systems}, 2023.

\bibitem[Zitkovich et~al.(2023)Zitkovich, Yu, Xu, Xu, Xiao, Xia, Wu, Wohlhart, Welker, Wahid, Vuong, Vanhoucke, Tran, Soricut, Singh, Singh, Sermanet, Sanketi, Salazar, Ryoo, Reymann, Rao, Pertsch, Mordatch, Michalewski, Lu, Levine, Lee, Lee, Leal, Kuang, Kalashnikov, Julian, Joshi, Irpan, Ichter, Hsu, Herzog, Hausman, Gopalakrishnan, Fu, Florence, Finn, Dubey, Driess, Ding, Choromanski, Chen, Chebotar, Carbajal, Brown, Brohan, Arenas, and Han]{zitkovich2023rt2}
Brianna Zitkovich, Tianhe Yu, Sichun Xu, Peng Xu, Ted Xiao, Fei Xia, Jialin Wu, Paul Wohlhart, Stefan Welker, Ayzaan Wahid, Quan Vuong, Vincent Vanhoucke, Huong Tran, Radu Soricut, Anikait Singh, Jaspiar Singh, Pierre Sermanet, Pannag~R. Sanketi, Grecia Salazar, Michael~S. Ryoo, Krista Reymann, Kanishka Rao, Karl Pertsch, Igor Mordatch, Henryk Michalewski, Yao Lu, Sergey Levine, Lisa Lee, Tsang-Wei~Edward Lee, Isabel Leal, Yuheng Kuang, Dmitry Kalashnikov, Ryan Julian, Nikhil~J. Joshi, Alex Irpan, Brian Ichter, Jasmine Hsu, Alexander Herzog, Karol Hausman, Keerthana Gopalakrishnan, Chuyuan Fu, Pete Florence, Chelsea Finn, Kumar~Avinava Dubey, Danny Driess, Tianli Ding, Krzysztof~Marcin Choromanski, Xi Chen, Yevgen Chebotar, Justice Carbajal, Noah Brown, Anthony Brohan, Montserrat~Gonzalez Arenas, and Kehang Han.
\newblock Rt-2: Vision-language-action models transfer web knowledge to robotic control.
\newblock In \emph{Proceedings of The 7th Conference on Robot Learning}, pages 2165--2183. PMLR, 2023.

\end{thebibliography}
\endgroup

\end{document}